\documentclass[10pt]{cai}
\begin{document}

\begin{center}
\title{Contextual Attention-Based Multimodal Fusion of LLM and CNN for Sentiment Analysis}
\maketitle

\thispagestyle{empty}

\begin{tabular}{cc}
Meriem Zerkouk\textsuperscript{\affilone,*}, Miloud Mihoubi\textsuperscript{\affilone}, Belkacem Chikhaoui\textsuperscript{\affilone} \\
{\small \textsuperscript{\affilone} Artificial Intelligence Institute, University of TELUQ, 5800, rue Saint-Denis,
Montreal, Quebec, H2S 3L5, Canada} \\
\end{tabular}
  
\emails{
  \textsuperscript{*}meriem.zerkouk@teluq.ca 
}
\vspace*{0.2in}
\end{center}

\begin{abstract}
This paper introduces a novel approach for multimodal sentiment analysis on social media, particularly in the context of natural disasters, where understanding public sentiment is crucial for effective crisis management. Unlike conventional methods that process text and image modalities separately, our approach seamlessly integrates Convolutional Neural Network (CNN) based image analysis with Large Language Model (LLM) based text processing, leveraging Generative Pre-trained Transformer (GPT) and prompt engineering to extract sentiment relevant features from the CrisisMMD dataset. To effectively model intermodal relationships, we introduce a contextual attention mechanism within the fusion process. Leveraging contextual-attention layers, this mechanism effectively captures intermodality interactions, enhancing the model's comprehension of complex relationships between textual and visual data. The deep neural network architecture of our model learns from these fused features, leading to improved accuracy compared to existing baselines.

Experimental results demonstrate significant advancements in classifying social media data into informative and noninformative categories across various natural disasters. Our model achieves a notable 2.43\% increase in accuracy and 5.18\% in F1-score, highlighting its efficacy in processing complex multimodal data. Beyond quantitative metrics, our approach provides deeper insight into the sentiments expressed during crises. The practical implications extend to real time disaster management, where enhanced sentiment analysis can optimize the accuracy of emergency interventions. By bridging the gap between multimodal analysis, LLM powered text understanding, and disaster response, our work presents a promising direction for Artificial Intelligence (AI) driven crisis management solutions.
\end{abstract}

\section{Introduction}

In recent years, the rapid growth of digital communication platforms has led to an unprecedented surge in multimodal data, encompassing text, images, and audio \cite{Mai2024DynamicGC}. While this wealth of information offers deep insights into human expression and interaction, it also presents significant challenges in analyzing and understanding the sentiments embedded within these diverse modalities. Sentiment analysis—the computational task of identifying attitudes, opinions, and emotions from text, images, audio, or their combination—has become a crucial area of research with wide-ranging applications, from market analysis to social media monitoring and beyond.

Traditionally, sentiment analysis has predominantly focused on uni-modal data, primarily textual content extracted from sources such as social media posts, product reviews, and news articles \cite{Milleville2023ExploringTP}. While text-based approaches have yielded valuable insights, they inherently overlook the nuanced information conveyed through other modalities, such as facial expressions, gestures, tone of voice, and contextually relevant visual cues. Traditional sentiment analysis methods suffer from one or more of the following limitations: 1) They often focus only on text, overlooking valuable insights from other modalities such as images and audio. 2) They struggle to capture the rich contextual information present in multimodal data. Sentiments expressed in text may be heavily influenced by accompanying visual or auditory cues, and analyzing each modality in isolation fails to account for these contextual nuances. 3) They often lack scalability and generalizability, particularly when applied to diverse datasets and domains. Uni-modal models trained on specific types of data may struggle to generalize to new contexts or modalities, requiring extensive retraining or fine-tuning for each new application. 4) They fail to capture the complex interplay between different modalities in conveying sentiment. For example, a positive sentiment expressed in text may be contradicted by negative facial expressions in an accompanying image, highlighting the importance of analyzing cross-modal dynamics. 

Recognizing these limitations, several research works have increasingly turned towards multimodal sentiment analysis, which integrates information from multiple modalities to achieve a more comprehensive understanding of sentiment \citep{Dong2023RecentAA,Chelabi2021ComparisonOD}. Data fusion in multimodal sentiment analysis leverages the complementary strengths of different modalities to enhance sentiment understanding. For instance, while text may provide explicit expressions of opinions and attitudes, visual cues such as facial expressions and body language can offer additional contextual information and emotional nuances that enrich the sentiment analysis process. Similarly, auditory signals, including intonation and vocal prosody, contribute valuable cues to infer underlying emotions and sentiments. By combining information from multiple modalities, the aim is to capitalize on the strengths of each modality while mitigating their individual limitations, thus enhancing the overall accuracy and robustness of sentiment analysis systems \citep{Zadeh2018MultimodalLA}. Furthermore, data fusion in multimodal sentiment analysis enables the exploration of synergistic effects that arise from the interaction between different modalities. By leveraging cross-modal correlations and dependencies, fusion techniques facilitate a more holistic interpretation of sentiment, enabling deeper insights into the underlying emotional states and attitudes expressed across diverse media types. Moreover, fusion approaches offer the flexibility to adapt to the dynamic and evolving nature of multimodal data sources, thereby enhancing the adaptability and generalizability of sentiment analysis models across various domains and application scenarios. Motivated by these challenges, we propose a novel approach that leverages the complementary strengths of Convolutional Neural Network (CNNs) and Large Language Model (LLMs), enhanced by contextual attention and prompt engineering. To address these limitations, we propose a novel multimodal sentiment analysis framework that directly overcomes the challenges identified above, notably by enabling fine-grained interaction modeling between modalities and long-range contextual understanding.  Our approach is distinguished by the following key contributions: Novel integration of LLM and CNN: We combine Generative Pre-trained Transformer (GPT) based text modeling with CNN-based image analysis to exploit the strengths of both modalities in sentiment representation.
Contextual Attention Mechanism for Multimodal Fusion: Unlike prior works that rely on basic concatenation or fixed fusion strategies, our model introduces a dynamic, contextual attention layer that enables the model to selectively focus on informative modality interactions. Prompt Engineering for Enhanced Text Understanding: The use of GPT prompts allows more semantically rich extraction of sentiment features from textual content, enhancing interpretability and depth. Our approach leverages a novel contextual attention mechanism to address Multimodal Sentiment Analysis (MSA) challenges, mitigating data redundancy, enabling effective fusion of heterogeneous modalities, and ensuring inter-modal integration. Additionally, our method contributes to: 1) enhanced precision and depth of sentiment and emotion identification in multimodal data; 2) practical implications for disaster management and response strategies through improved understanding of complex emotions during crises. Through these contributions, our proposed approach represents a significant advancement in the field of sentiment analysis, particularly in the context of multimodal data interpretation.  

The remainder of this paper is structured as follows: Section 2 reviews related work. Section 3 outlines our proposed approach, including feature extraction, fusion, and sentiment classification. Section 4 details the dataset and experimental setup, while Section 5 presents performance evaluation. Section 6 discusses the ablation study, and Section 7 concludes with key findings and future directions.

\section{Related work}

Sentiment analysis and emotion recognition are emerging trends in social media analysis. They provide valuable insights into the behavior, preferences, and content of social media users \citep{Zerkouk2025PredictingOE,Tshimula2020OnPB}. Although much of the previous research has concentrated on text-based sentiment analysis using data from tweets, text sentiment analysis has changed significantly over time. It has transitioned from conventional rule-based techniques to sophisticated machine learning, deep learning techniques, and pre-trained language models \citep{Mai2024DynamicGC}.
Exploring sentiment analysis through images has also made significant advances, evolving from fundamental techniques based on basic image attributes to advanced deep learning approaches that integrate complex visual data. However, integrating text with images for sentiment analysis poses a complex challenge.  

Yue et al. \citep{Yue2020SentimentAU} introduced a model that  combines the strengths of CNNs for feature extraction with the capacity of Bi-LSTMs to understand the bidirectional contextual dependencies in text, especially in short text sequences. Their findings indicated that such combined models have superior performance compared to neural networks with a singular architecture when analyzing brief texts. Their results demonstrated that hybrid network models outperformed single-structure neural networks in the context of short texts. 
Abavisani et al. \citep{Abavisani2020MultimodalCO} introduced a multimodal sentiment analysis method using Bidirectional Encoder Representations from Transformers (BERT) to capture long-distance dependencies in text. This approach highlighted the importance of viewing image captioning as a multistep cognitive process rather than a simple translation task. Wang et al. \citep{s23052679} proposed Multimodal Sentiment Analysis Representations Learning via Contrastive Learning with Condense Attention Fusion. This work made a pivotal move towards using supervised contrastive learning in multimodal sentiment analysis. Proposing the MLFC module showcased an innovative approach to managing redundant information across modalities, which is a recurrent challenge in the field. Kumar et al. et al. \citep{Kumar2020ADM} introduced  a method combining text and image data from Twitter to identify content relevant to disaster response. The approach leverages a Long Short-Term Memory (LSTM) network and a VGG-16 network to process the multimodal data, leading to a notable performance improvement.  Ofli et al. \citep{Ofli2020AnalysisOS} introduced a deep learning approach that combines VGG16 for image processing and a mix of word2vec and CNN for text analysis into a shared feature space, leading to a unified classification via a feedforward neural network. This multimodal method notably enhances disaster-related information analysis on social media, yielding a 78.4\% accuracy and outperforming single-modality models. 

Yang et al. \cite{Yang2020ImageTextME} proposed the Multimodal Variational Autoencoder Network (MVAN), a multimodal sentiment analysis method that utilizes memory network modules and multi-view attention mechanisms to iteratively extract semantic features from both images and text. While effective, MVAN primarily focuses on feature extraction but does not account for the dynamic interactions between modalities. 
Shetty et al. \citep{Shetty2024DisasterAF} introduced a middle fusion paradigm combining cross-modal attention and self-attention to integrate textual and visual features for disaster-related event detection. The approach outperformed  early and late fusion methods, achieving superior accuracy in informativeness and disaster type classification on the CrisisMMD dataset, with a 2\% improvement over prior multimodal techniques. 

The aforementioned approaches present certain drawbacks related to intermodal and intramodal relationships and feature prioritization: 
Understanding the context is challenging because  existing models examine text and image modalities separately without considering the context in which they occur. Current multimodal sentiment analysis methods have various drawbacks. They often rely on essential fusion techniques like concatenation because they struggle to handle long-distance relationships in text sequences and effectively fuse text and image modalities. Furthermore, these models have difficulty determining and prioritizing the most relevant features from each modality for sentiment analysis.

Our approach overcomes these limitations through the following methodology:  - Understanding Intermodal and Intramodal Relationships where our model delves into deciphering the complex relationships between different modalities (intermodal) and within the same modality (intramodal).  - GPT’s ability to process long-range dependencies, combined with prompt engineering, enhances contextual understanding for more precise feature extraction. By combining the strengths of CNNs and LLMs, key sentiment features are prioritized, leading to improved sentiment prediction. Ensuring effective multimodal fusion enables the capture of complex interactions between text and image modalities.
By addressing these challenges, our framework significantly enhances the depth, accuracy, and interpretability of multimodal sentiment analysis, ensuring a comprehensive understanding of features and their complex interconnections.

\section{Proposed Approach}
The goal of multimodal data fusion for sentiment analysis is to identify the strength of emotions expressed in a tweet by analyzing multiple modalities. Our multimodal sentiment analysis model integrates textual, visual, and contextual attention data to predict sentiments. The core equation of our model is described as follows:
\begin{equation}
    \hat{s} = f(\varphi(X, Y, C); \Theta)
\end{equation}
where:
\begin{itemize}
   \item $\hat{s}$ represents the predicted sentiment, which can be a definite value (positive, negative or neutral). 
    \item $f$ is the classification  function, potentially realized through a neural network, which maps integrated features to sentiment predictions.
    \item $\varphi(X, Y, C)$ is the multimodal feature integration function that combines features from text $X$, image $Y$, and contextual information $C$ into a unified representation.
    \item $\Theta$ encompasses all model parameters, including those specific to feature extraction, feature integration, and sentiment prediction function.
\end{itemize}

\begin{figure}[h]
  \centering
  \includegraphics[width=\linewidth]{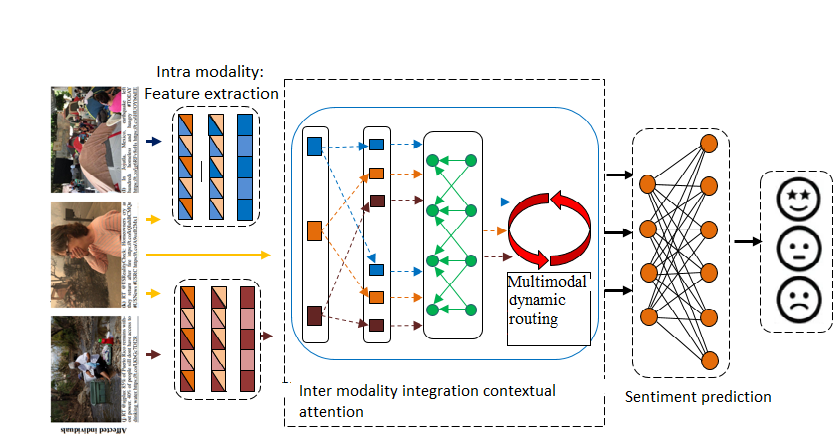}
    \caption{Proposed multimodal sentiment analysis architecture.}
    \label{fig:approach}
    \end{figure}

This formula succinctly captures the operation of our proposed model by highlighting the process of integrating different modalities of data to enhance sentiment analysis, particularly in complex scenarios like disaster response on social media platforms.
The architecture described in Figure~\ref{fig:approach} is designed explicitly for sentiment analysis, focusing on inputs that combine images and text, such as tweets that may or may not be informative. It incorporates three main components to process these inputs effectively. 
- The first module: Intra-Modal Discriminative Feature Extraction, is responsible for analyzing and extracting relevant features within each modality (image and text). 
- The second module: Inter-Modal Integration with Contextual Attention modeling, integrates these features by employing contextual attention to enhance the analysis by focusing on the most relevant information across both modalities. Together, these components enable a comprehensive understanding of sentiment in complex image-text inputs.
- The third module : Sentiment prediction module employs a Recurrent Neural Network (RNN) to integrate the features obtained from the Inter-Modal Integration with Contextual Attention module. It provides a prediction of the sentiment, which can be classified into categories such as positive, negative, or neutral, as illustrated in Figure ~\ref{fig:approach}.

\subsection{Intra-Modal Discriminative Feature Extraction}

Intra-modal discriminative feature extraction isolates and enhances meaningful features within each modality (text or image) independently, preserving their integrity before inter-modal integration or analysis.

\subsubsection{Text Feature Extraction}

An LLM-powered approach is used to extract text features from tweets, with the GPT model enhancing representation through its transformer architecture. This allows it to capture long range dependencies in text an essential skill for analyzing sentiment across a series of tweets. The average sequence length is ~23 words per tweet, aligning with the character limitations of Twitter. The process starts with preprocessing the tweets, which includes cleaning and normalizing the text to ensure high quality input before it goes into the model. To optimize performance, we employ various GPT model variants, incorporating prompt engineering to refine text processing. The baseline GPT model applies the standard architecture without modifications or prompts, serving as a reference for evaluating advanced configurations.

In the GPT model augmented with a prompt, the prompt \(P_r\) is combined with the original text \(t'\) to create an enriched input sequence \([P_r, t']\), which is then processed by the GPT model:
\begin{equation}
E = \{ f([P_r, t']) \mid t' \in T' \}.
\end{equation}
This approach utilizes the prompt's capacity to guide the model's attention toward specific contextual aspects of the text, thereby improving the significance and precision of the features extracted.

Building on this, the GPT model with prompt and variant text introduces multiple text variants in conjunction with the prompt. The input sequence now includes several variations of the original tweet text \(t_i'\), combined with the prompt \(P_r\), enriching the context and improving the model's ability to generalize across different textual representations:
\begin{equation}
E = \{ f([P_r, t_i']) \mid t_i' \in T' \times V \},
\end{equation}
where \(V\) represents a set of varied text sequences. These variations are generated to capture different potential interpretations or contexts of the original text.

Finally, the GPT model with prompt and fine-tuning further enhances the model by fine-tuning it on a task-specific dataset while also incorporating a prompt. This fine-tuning process involves adjusting the model's parameters to better match the specific details of the sentiment analysis task, leading to more accurate and reliable feature extraction.

Let \( f \) denote the initialized GPT model, which takes a preprocessed text sequence \( X(t) \) and maps it to a feature vector \( F(t) \).
This function can be formulated as: 
\begin{equation}
f : X(t) \rightarrow F(t).
\end{equation}

Here, \( F(t) \) represents the feature vector resulting from the application of the GPT model to the text sequence \( X(t) \). The feature vector \( F(t) \) \textbf{is} then used for subsequent tasks such as fusion and sentiment analysis.

\subsubsection{Image feature extraction}
 
ResNet50's architecture, known for its deep residual learning, enables the efficient training of deeper networks. Its skip connections, or shortcuts, allow the network to bypass specific layers, effectively addressing the vanishing gradient problem that often occurs in deep networks.

Initialize a ResNet50 model with a specified number of output classes and without pretrained weights. Mathematically, the CNN is a function \( \text{C} \) that maps an input image \( Q \) to a prediction vector \( \hat{y} \):
\begin{equation}
\text{C} : Q \rightarrow \hat{y}
\end{equation}

The feature vector used for analysis is extracted from the penultimate layer of ResNet50. This vector encapsulates critical characteristics of the input image, making it suitable for various analytical applications. It can be used directly in classification tasks or integrated into further multimodal analysis processes, providing flexibility in its application. Typically, the feature vector is 2048 dimensional and is incorporated into the multimodal interaction layer.

\subsection{Inter-Modal Integration with Contextual Attention}
Our multimodal fusion module integrates textual features from GPT and visual features from ResNet50, utilizing their complementary strengths for enhanced sentiment and emotion classification. To effectively integrate these diverse modalities, our model architecture incorporates context functions and dynamic routing \citep{Sabour2017DynamicRB}. Our approach is designed specifically for sentiment analysis, using contextual attention and dynamic routing to intelligently combine text and visual data. In contrast, models like CLIP \citep{Radford2021LearningTV} and Flamingo \citep{Alayrac2022FlamingoAV} rely on static multimodal alignment and aren't optimized for sentiment detection.

We chose to fine-tune GPT and integrate an optimized CNN to ensure both interpretability and real-time efficiency. While Flamingo and CLIP are powerful, they lack domain-specific adaptation and cannot dynamically adjust the importance of each modality based on context. The context functions dynamically capture and prioritize relevant features through a contextual attention mechanism, while dynamic routing optimizes data flow through iterative updates, refining feature representations over multiple steps to improve alignment and robustness. This iterative process helps reduce redundancy, enhance accuracy, and improve computational efficiency.

Dynamic routing involves iterative updates, where the model continuously refines cross-modal compatibility by evaluating the alignment between textual and visual features and adjusting their importance dynamically. By capturing context at multiple levels, our architecture ensures fine-grained fusion, refining feature vectors through contextual attention to produce a more structured and interpretable multimodal representation.

Feature extraction for text and image modalities begins with extracting the raw feature sets, denoted as $\mathbf{F}^{(T)}$ for text and $\mathbf{F}^{(I)}$ for images. The extracted feature vectors are:

\begin{equation} \mathbf{X}^{(T)} = f_{\text{extract}}^{(T)}(\mathbf{F}^{(T)}) \end{equation} \begin{equation} \mathbf{X}^{(I)} = f_{\text{extract}}^{(I)}(\mathbf{F}^{(I)}) \end{equation}

After extraction, these features are combined into a joint latent representation $Z$, which captures interactions across multiple abstraction levels, ensuring a multi-level contextual understanding of sentiment:

\begin{equation} Z = g(\mathbf{X}^{(T)}, \mathbf{X}^{(I)}) \end{equation}

Finally, the joint latent representation is mapped to the sentiment space $Y$, which could be a predicted sentiment label or score:

\begin{equation} Y = h(Z) \end{equation}

The learning process aims to minimize the average loss over a dataset $\mathcal{D}$ using the following objective:

\begin{equation} \min_{g, h} \frac{1}{N} \sum_{i=1}^{N} \mathcal{L}(h(g(\mathbf{x}^{(i)})), y^{(i)}) \end{equation}

This is a multiclass classification problem with $C$ classes, where $y_{o,c}$ is a binary indicator (0 or 1) that denotes whether class label $c$ is the correct classification for observation $o$, and $h_{o,c}$ represents the predicted probability that observation $o$ belongs to class $c$. Let $\mathbf{f}{\text{text}}$ and $\mathbf{f}{\text{image}}$ be the attention-weighted feature vectors for text and image modalities, respectively. The concatenated feature vector $\mathbf{f}$ can be represented as:

\begin{equation} \mathbf{f} = [ \mathbf{f}{\text{text}} ||  \mathbf{f}{\text{image}} ] \end{equation}

Here, $\|$ denotes the fusion operation. This combined feature vector $\mathbf{f}$ is formed by fusing contextual attentive uni-modal features to create a bimodal representation that enhances cross-modal interpretability. The iterative update mechanism within dynamic routing refines feature interactions, and the multi-level contextual approach ensures the model effectively captures both low- and high-level relationships between modalities. This context-aware fusion process leads to more accurate and robust sentiment classification.

\section{Experiments Result and Discussion}

\subsection{Dataset}

Our study employs the "CrisisMMD" dataset \cite{Cresci2015CrisisMD}, which includes text and image data from seven major natural disasters in 2017, such as Hurricane Harvey and the Mexico Earthquakes. The dataset was divided into training, validation, and test sets, with text and image instances labeled as informative or non-informative, comprising 11,400 text instances and 12,708 image instances. Comprehensive cleaning processes were applied to both modalities to reduce false positives and ensure data reliability. Additionally, image data was augmented through techniques like rotation, flipping, and shifting.  This dataset offers a strong basis for analyzing and improving multimodal fusion methods.

\subsection {Performance Evaluation }\label{sec2}
Our approach to studying multimodal sentiment analysis takes advantage of the semantic features from text and the features from images found in CrisisMMD. We categorized each tweet in the dataset as informative or non-informative in CrisisMMD.  Our model training and evaluation were conducted on this enriched dataset across the different distinct disaster events, aiming to advance the multimodal data fusion in the context of disaster-related information on social media.

\subsubsection{Parameter Tuning}
During the experiments, we explored how different learning rates and batch sizes impacted the model's performance. The batch sizes used were 32 and 64, while the learning rates tested were 0.05, 0.01, 0.005, and 0.001. The final model was constructed using the parameters that provided the best results. Table \ref{tab:parameter_settings} presents the final parameter configurations for our proposed approach:

\begin{table}[ht]
    \caption{Experimental Parameter Settings}
    \label{tab:parameter_settings}
    \centering
    \begin{tabular}{|l|c|}
       \textbf{Parameters} & \textbf{Values} \\ \hline
        GPT unit & 30 \\ \hline
        Fully connected unit & 100 \\ \hline
        Dropout & 0.5 \\ \hline
        Learning rate & 0.001 \\ \hline
        Batch size & 32 \\ \hline
        Number of iterations & 100 \\ \hline
        Optimization function & Adam \\ \hline
        Loss function & Binary cross-entropy \\ 
    \end{tabular}
\end{table}

\subsubsection {Performance Evaluation of GPT Model for Text Modality}\label{sec2}

Our study evaluates multiple models for text classification on the CrisisMMD dataset, comparing GPT and BERT against traditional machine learning algorithms such as Naive Bayes, SVM, and Random Forest. The evaluation is based on key metrics such as accuracy and F1-score, as shown in Table \ref{tab:text_analysis_performance}. To mitigate dataset biases and improve efficiency, we apply data balancing techniques and fine-tune hyperparameters for the GPT model. When comparing GPT and BERT in classifying tweets as Informative vs. Non-Informative, GPT consistently outperforms BERT across multiple key metrics. Specifically, GPT achieves an F1-score of 90.40\%, an accuracy of 84.91\%, and a precision of 90.76\%, while BERT reaches an F1-score of 89.00\%, an accuracy of 88.89\%, and a precision of 89.00\%. These results highlight GPT’s superior ability to capture contextual nuances, making it more effective in tasks requiring comprehensive instance identification. Meanwhile, BERT’s slightly higher accuracy suggests its advantage in minimizing false positives, making it suitable for applications where misclassification penalties are high.  A key advantage of GPT lies in its flexibility through prompt engineering, which allows for better adaptation to specific classification tasks compared to BERT’s static embeddings. This adaptability enables GPT to capture nuanced distinctions in crisis-related tweets more effectively, reinforcing its suitability for real-world disaster response scenarios.

\begin{table}[ht]
\centering
\caption{Performance Metrics of Different Models for Text Analysis.}
\label{tab:text_analysis_performance}
\begin{tabular}{|p{2.6cm}|p{1.4cm}|p{1.2cm}|p{1.2cm}|}

\textbf{Models} & \textbf{F1-score} & \textbf{Accuracy} & \textbf{Precision} \\ \hline
RF         & 88.00 & 89.67 & 88.00 \\ \hline
SVM        & 89.00 & 89.32 & 89.00 \\ \hline
Gaussian NB & 74.00 & 75.00 & 80.00 \\ \hline
GLOVE      & 88.00 & 88.07 & 88.00 \\ \hline
Word2Vec   & 89.00 & 88.75 & 89.00 \\ \hline
BERT       & 89.00 & 88.89 & 89.00 \\ \hline
GPT        & 90.40 & 84.91 & 90.76 \\ 
\end{tabular}
\end{table}

\subsubsection{Performance Evaluation of CNN models for image modality}\label{sec2}
ResNet50 has achieved significant results on the disaster dataset, demonstrating its effectiveness in classifying and distinguishing between images from Informative and Non-Informative tweets. In classifying images from Informative tweets, ResNet50 has shown outstanding performance, as detailed in Table \ref{tab:model_performance-img}, highlighting its capability to accurately distinguish between Informative and Non-Informative categories. These results are particularly significant given the critical role of accurate classification in humanitarian response efforts, where the precise identification of relevant information can directly impact the effectiveness of disaster response. Furthermore, ResNet50 has demonstrated a high level of accuracy, achieving 88.56\% in this task, underscoring its robust feature extraction and classification capabilities.

\begin{table}[ht]
\centering
\caption{Performance Metrics of Different Models for Image Analysis.}
\label{tab:model_performance-img}
\begin{tabular}{|p{2.7cm}|p{1.4cm}|p{1.2cm}|p{1cm}|}
\textbf{Models} & \textbf{F1-score} & \textbf{Accuracy} & \textbf{Precision} \\ \hline
VGG16         & 71.00 & 71.04 & 71.00 \\ \hline
VGG19         & 74.00 & 73.96 & 74.00 \\ \hline
InceptionV3   & 80.00 & 80.51 & 80.00 \\ \hline
DenseNet121   & 82.00 & 81.59 & 82.00 \\ \hline
EfficientNetV2 & 77.00 & 77.10 & 77.00 \\ \hline
ResNet50      & 88.00 & 88.56 & 88.00 \\ 
\end{tabular}
\end{table}

\subsubsection {Performance evaluation of Fusion Text and Image Modalities }\label{sec2}
It is crucial to carefully select the most appropriate model for our problem, prioritizing those that minimize information loss. To this end, Table \ref{tab:performance} provides detailed metrics on accuracy, recall, and F1-score. Our approach achieved a notable accuracy of 93.75\%, coupled with a recall of 93.75\% and an F1-score of 96.77\%. These results reflect a 2.43\% increase in accuracy and 5.18\% in F1-score compared to our baseline models, underscoring the advantage of integrating text and image modalities for more comprehensive sentiment analysis over using uni-modal data alone.  Beyond our proposed model, various fusion strategies were explored across different dataset categories, with a clear distinction between intra- and inter-modal data interactions as outlined in the methodology. These strategies were instrumental in achieving superior performance metrics.

\begin{table}[h!] 
\centering 
\caption{Performance metrics of our model on Informative vs Non-Informative categories.}
\scriptsize 
\begin{tabular}{|p{3.5cm}|p{1.5cm}|p{1.5cm}|p{1.5cm}|} 

\textbf{Models} & \textbf{Accuracy (\%)} & \textbf{Recall (\%)} & \textbf{F1-Score (\%)} \\ 
\hline 
LLM+CNN (Ours) & \textbf{93.75} & \textbf{93.75} & \textbf{96.77} \\ \hline
Shetty et al.\citep{Shetty2024DisasterAF} & 91.53 & 92.00 & 92.00 \\ \hline
Kota et al.\citep{Kota2022MultimodalCO} & 89.63 & 89.59 & 89.59 \\ \hline
Abavisani et al.\citep{Abavisani2020MultimodalCO} & 89.33 & -- & 89.00 \\ \hline
Sirbu et al.\citep{Sirbu2022MultimodalSL} & -- & 91.00 & 91.00 \\ \hline
Zou et al.\citep{Zou2021DisasterIC} & 87.60 & 88.00 & 88.00 \\ \hline
Ofli et al.\citep{Ofli2020AnalysisOS} & 84.40 & 84.40 & 84.40 \\ 
\end{tabular} 
\label{tab:performance}
\end{table}

The efficacy of these fusion strategies was rigorously evaluated, and our model outperformed traditional methods employing other multimodal fusion strategies. The results demonstrate that effective feature integration can reveal complex, hidden associations between modalities that are often oversimplified or overlooked by other techniques. This supports our hypothesis that the depth of contextual understanding is significantly enhanced when features interact dynamically.

Our multimodal data fusion approach benefited not just from the combination of textual and visual cues, but from the contextualized fusion, which facilitated the nuanced interaction between these modalities. The addition of contextual attention provided the granularity needed for a detailed sentiment analysis, contributing to the improved accuracy and robustness of our model.

\section{Ablation Study}
To evaluate the contribution of each component in our model, we conducted an ablation study on the multimodal fusion approach using the CrisisMMD dataset. This study involved a series of experiments assessing the effectiveness of individual model components, particularly text and image feature extraction techniques, to determine their impact on multimodal sentiment analysis.

We tested the following model configurations:

\begin{itemize}
    \item \textbf{Text Only (Simple GPT)}: Evaluates only the text modality using the base GPT model.
    \item \textbf{Text Only (GPT, Single Prompt)}: Incorporates a single prompt to direct the model’s focus on specific contextual aspects, refining text representation.
    \item \textbf{Text Only (GPT, Prompt with Varied Text)}: Uses multiple text variations to improve generalization by exposing the model to diverse contextual interpretations.
    \item \textbf{Text Only (GPT, Prompt and Fine-Tuning)}: Fine-tunes the GPT model on a task-specific dataset while utilizing a prompt to optimize sentiment analysis performance.
    \item \textbf{Image Only (CNN)}: Evaluates the model using only the image modality without textual context.
    \item \textbf{Contextual-Attention (Ours)}: Integrates textual and visual modalities through a contextual attention mechanism, dynamically weighting features based on relevance.
\end{itemize}

\begin{table}[ht]
\centering
\caption{Ablation Study: Performance of Different Configurations on CrisisMMD Dataset}
\label{tab:ablation_experiments}
\begin{tabular}{|p{4.5cm}|p{1.4cm}|p{1.2cm}|}

\textbf{Models} & \textbf{F1-Score} & \textbf{Accuracy} \\ \hline
(Simple GPT) & 90.40 & 84.91 \\ \hline
(GPT, single prompt) & 92.43 & 87.74 \\ \hline
(GPT, prompt with varied text) & 92.70 & 88.36 \\ \hline
(GPT, prompt and fine-tuning) & 93.10 & 88.99 \\ \hline
Image Only (CNN) & 81.00 & 88.56 \\ \hline
Contextual-Attention (Ours) & \textbf{96.77} & \textbf{93.75} \\ 

\end{tabular}
\end{table}

When analyzing a tweet like "The flood is rising, people are trapped," task-specific prompts refine text feature extraction by guiding the LLM to assess urgency, classify event type, identify affected locations, and analyze sentiment polarity. This structured approach enhances contextual representation, improving classification accuracy and multimodal fusion effectiveness.

\subsection{Discussion}

The ablation study reveals key insights into the model's performance across different configurations. The LLM-based text models demonstrate strong effectiveness in sentiment analysis, with the simple GPT model achieving an F1-score of 90.40\% and an accuracy of 84.91\%. Introducing a prompt further enhances performance, with the best LLM-based configuration (GPT with prompt and fine-tuning) achieving an F1-score of 93.10\% and an accuracy of 88.99\%. This underscores the importance of prompt engineering and fine-tuning in optimizing LLM performance for sentiment analysis.

The CNN-based image-only model, on the other hand, achieves an F1-score of 81.00\% and an accuracy of 88.56\%, highlighting the limitations of relying solely on visual features due to the lack of textual context.

Our proposed model, which combines text and image features through a contextual attention mechanism, achieves the highest performance with an F1-score of 96.77\% and an accuracy of 93.75\%. These results validate the effectiveness of multimodal fusion, where contextual attention dynamically balances textual and visual modalities, significantly improving sentiment analysis performance.

Since 85\% of CrisisMMD posts contain both text and images, multimodal fusion is crucial for effective sentiment classification. Our model leverages contextual attention to dynamically adapt, utilizing GPT for text or ResNet50 for images, ensuring robustness even in unimodal scenarios, as demonstrated in our ablation study.

\section{Conclusion}\label{sec4}

This paper presents a novel multimodal data fusion approach that addresses critical challenges in sentiment analysis across diverse modalities. The central difficulty lies in designing models capable of effectively analyzing and integrating heterogeneous data sources. Our methodology leverages a dynamic enhancement module to address both intramodal and intermodal complexities, particularly in text and image processing. In the preprocessing phase, we utilize LLMs, specifically the GPT model enhanced with prompt engineering for textual analysis, and CNNs for image processing. This combination enables the extraction of deep, sentiment-relevant features from both modalities, enhancing the model’s understanding of complex emotional contexts, especially in disaster-related scenarios. Furthermore, the integration of a contextual attention mechanism refines the fusion process by capturing intermodal dependencies, reducing redundancy, and significantly improving overall performance. Our model demonstrates a notable improvement over the best existing baseline, achieving a 2.43\% increase in accuracy and a 5.18\% gain in F1-score, underscoring its effectiveness for multimodal sentiment classification. Future work will extend this framework to include additional modalities such as audio and video, refine the contextual attention mechanism, and explore larger scale datasets to address increasingly complex sentiment analysis tasks. Although this study focuses on CrisisMMD for domain specific evaluation, future work will aim to assess the model’s robustness and generalizability across a broader range of multimodal datasets and domains.


\end{document}